\documentclass{article}
\usepackage{ijcai23}

\usepackage{times}
\usepackage{soul}
\usepackage{url}
\usepackage[hidelinks]{hyperref}
\usepackage[utf8]{inputenc}
\usepackage[small]{caption}
\usepackage{graphicx}
\usepackage{amsmath}
\usepackage{amsthm}
\usepackage{booktabs}
\usepackage{algorithm}
\usepackage{algorithmic}
\usepackage[switch]{lineno}

\usepackage{subfigure} 
\usepackage{pifont}
\usepackage{threeparttable}
\usepackage{amsfonts}

\title{Scene Style Text Editing\footnote{This work has been submitted to the IEEE for possible publication. Copyright may be transferred without notice, after which this version may no longer be accessible.}}

\author{
Tonghua Su$^1$\and
Fuxiang Yang$^1$\and
Xiang Zhou$^1$\and
Donglin Di$^2$ \and
Zhongjie Wang$^1$ \and
Songze Li$^1$
\affiliations
$^1$Harbin Institute of Technology\\
$^2$Tsinghua University\\
\emails
\{thsu, rainy\}@hit.edu.cn,
hityangfx@foxmail.com,
\{zx543977392, donglin.ddl\}@gmail.com,
lisongze92@163.com
}

\begin{document}

\maketitle

\begin{abstract}
In this work, we propose a task called ``Scene Style Text Editing (SSTE)'', changing the text content as well as the text style of the source image while keeping the original text scene.
Existing methods neglect to fine-grained adjust the style of the foreground text, such as its rotation angle, color, and font type.
To tackle this task, we propose a quadruple framework named ``QuadNet'' to embed and adjust foreground text styles in the latent feature space.
Specifically, QuadNet consists of four parts, namely background inpainting, style encoder, content encoder, and fusion generator.
The background inpainting erases the source text content and recovers the appropriate background with a highly authentic texture.
The style encoder extracts the style embedding of the foreground text.
The content encoder provides target text representations in the latent feature space to implement the content edits. 
The fusion generator combines the information yielded from the mentioned parts and generates the rendered text images. 
Practically, our method is capable of performing promisingly on real-world datasets with merely string-level annotation. 
To the best of our knowledge, our work is the first to finely manipulate the foreground text content and style by deeply semantic editing in the latent feature space. 
Extensive experiments demonstrate that QuadNet has the ability to generate photo-realistic foreground text and avoid source text shadows in real-world scenes when editing text content.

\end{abstract}

\begin{figure}[t]
    \centering 

\subfigure[Original]{
\centering
 \includegraphics[width=0.8\linewidth]{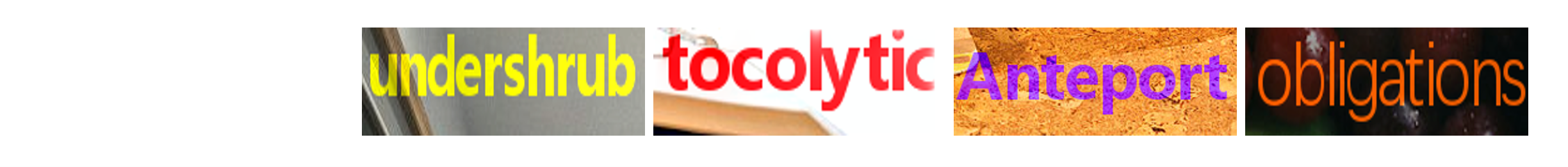}
}

\subfigure[SeFa]{
\centering
 \includegraphics[width=0.8\linewidth]{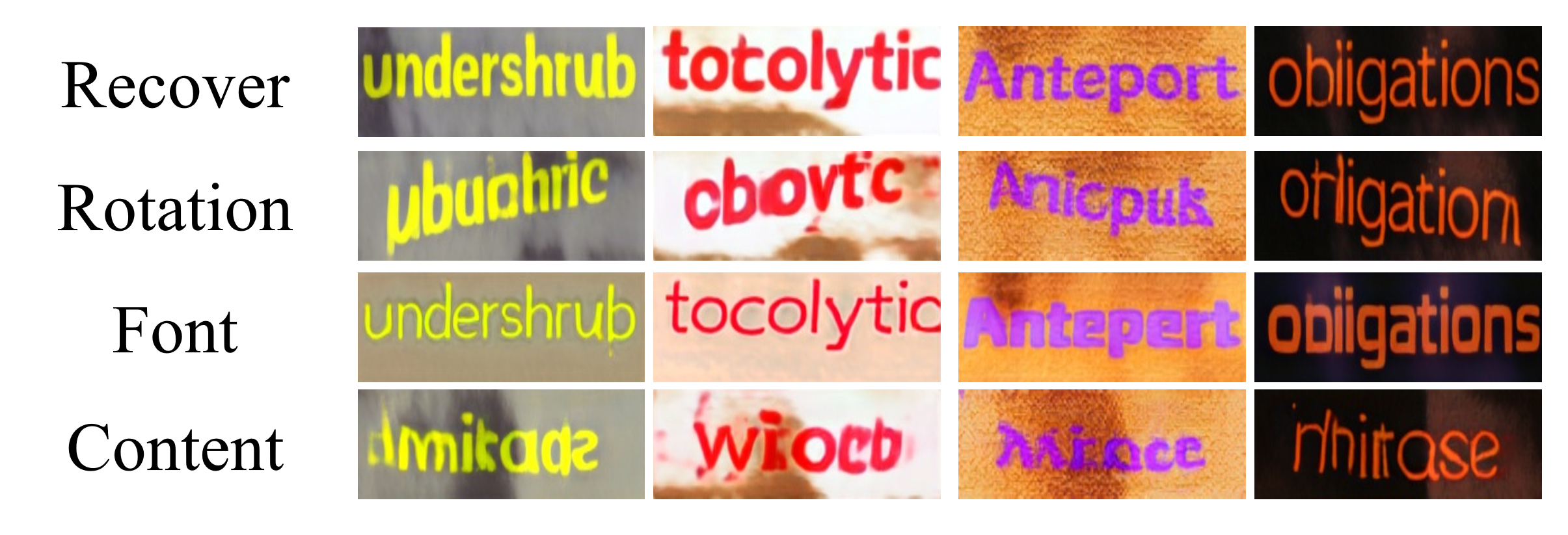}
}

\subfigure[QuadNet]{
\centering
 \includegraphics[width=0.8\linewidth]{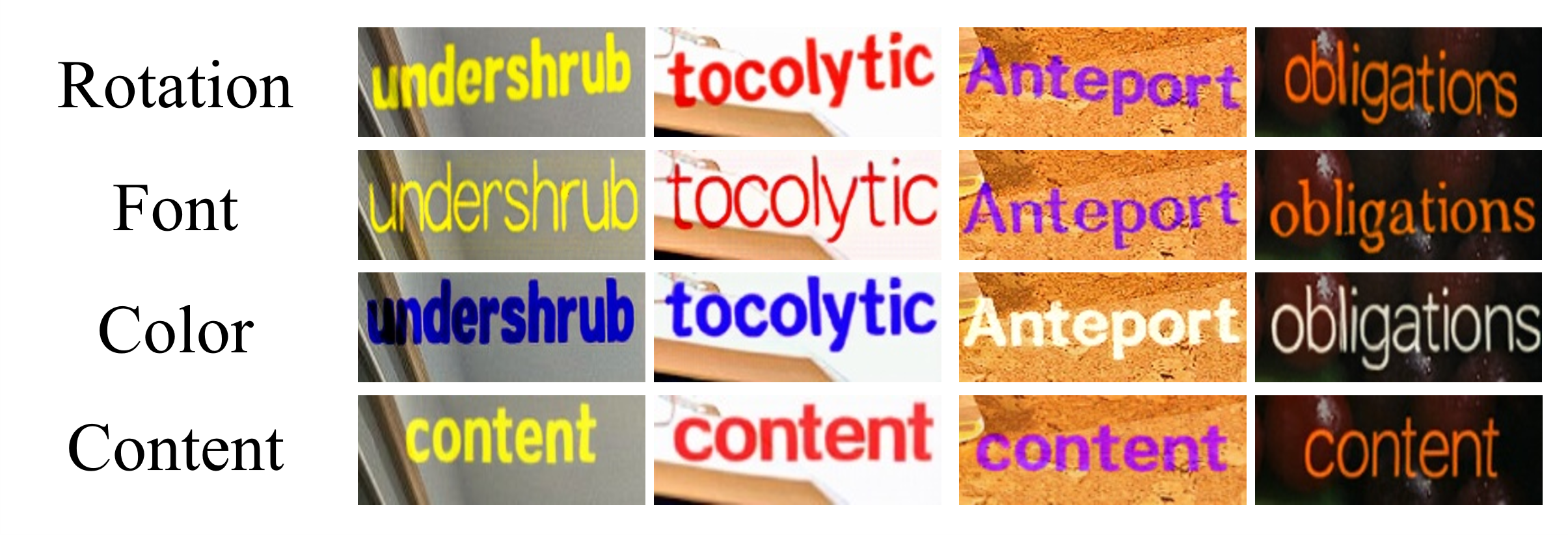}
}
    \caption{Qualitative comparison between SeFa.
    (a) Original Image.
    (b) SeFa: first reconstruct the image (with character errors, such as ``tocolytic" reconstructed to ``totclytic"), then modify the text rotation (text content is unreadable), font (thin or bold, but the background is also modified), content (unreadable and cannot be modified according to conditions). 
    (c) QuadNet: edit the text rotation angle (clockwise or counterclockwise), font (thinning or serif), color (blue or white) and content (can be modified conditionally).
    }
    \label{fig.sefa_vs_our}
\end{figure}
\begin{figure*}[t]
    \centering
    \includegraphics[width=0.94\textwidth]{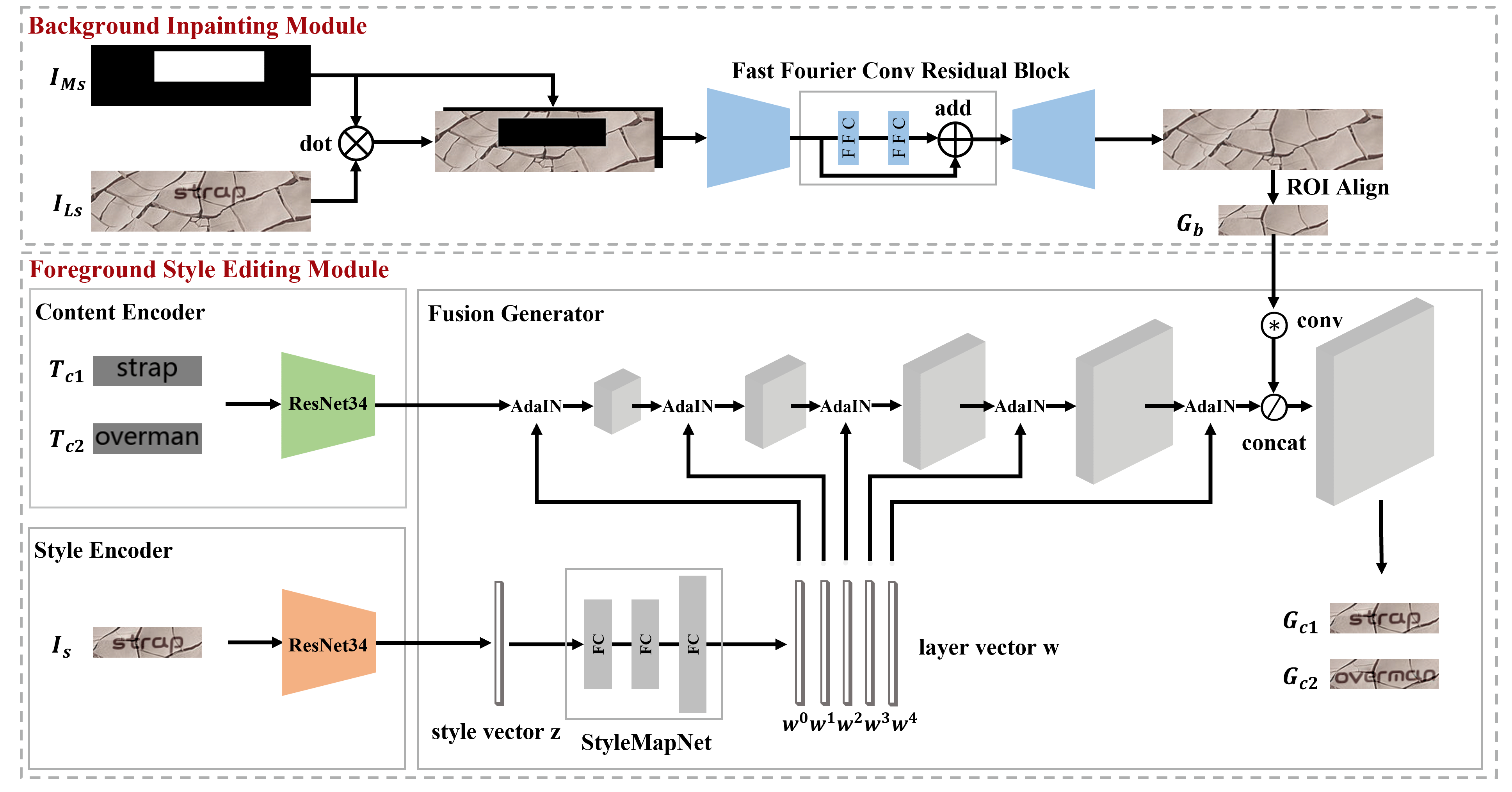}
    \caption{\label{fig:architecture}  The overall structure of QuadNet. The network consists of four parts, including background inpainting, style encoder, content encoder and fusion generator, which can be arranged into background inpainting module and foreground style editing module.
    }
    \end{figure*}

\section{Introduction}

Many designers have ever encountered such scenarios during their design works: 
1) need to rotate the text in a natural scene image a bit and change the font from red ``Arial'' to white ``Arial Black'' at the same time;
2) need to design a bundle of foreground texts with distinct style in a poster for customers to choose;
3) need to change the content of the foreground text in the scene image while keeping the original text styles.
The three cases above are interesting yet challenging. 
We should perform editing text content or fine-grained adjustment of the style of foreground text while recovering the complex background texture.

We call this task as \textbf{Scene Style Text Editing} (SSTE), which enables you to change the text style as you wish, such as text rotation angle, font type, text color, besides text content. 
Previously, scene text editing (STE) ~\cite{wu2019editing,yang2020swaptext,roy2020stefann,dendorfer2021mg,Luo2022SimAN} can only modify its text content. 
Recent emerging technology for editing faces opens new insights for the SSTE task. 
Some pioneer works ~\cite{ling2021editgan,shen2020interpreting,shen2021closed,Cherepkov_2021_CVPR} embed images into the GAN's latent space firstly. 
Then modifications occur in the latent space, and the desired images can generate through the GAN generator. 

SeFa ~\cite{shen2021closed} is a relatively advanced unsupervised semantic editing method for faces.
We modify SeFa and apply it to SSTE, uncovering some meaningful editing semantics such as rotating text, modifying text font or content, as shown in Figure \ref{fig.sefa_vs_our} (b). 
However, there are several issues with this approach. 
Rotating the text changes its characters, altering the font also affects the background, and the modified text content becomes unreadable. 
In contrast, as shown in Figure \ref{fig.sefa_vs_our} (c), our QuadNet produces clear and coherent results.
Our method separates the foreground and background, and decouples the style of the foreground text from its content. 
This allows us to effectively rotate the text, modify the font and color, without changing the text content or background texture.
We can also change the text content while maintaining the text style.

In this work, we propose a novel scene style text editing model named QuadNet, as shown in the Figure \ref{fig:architecture}. 
Our model includes four parts: background inpainting, style encoder, content encoder and fusion generator. 
They can be arranged into two modules in phases: background inpainting module and foreground style editing module. 
The former firstly discards the foreground text and recover the background texture, then use the region of interest (RoI) align operator to crop the background texture. 
The latter: 1) extracts the style embedding and content representations from the foreground style text and target content image; 2) decouples the style attributes, and fuses it with background texture and text content representations to generate image.
Our contributions are summarized as follows:
    \begin{itemize}
    \item To our knowledge, QuadNet is the first attempt to perform fine-grained adjustment of the style of foreground text in SSTE task by semantic editing in latent space. 
    
    \item  QuadNet supports editing of both text content and text style in latent space. 
    Our method firstly separates the background texture and foreground content/style, making the latent space easier to encode the foreground text. 
    Next, our method decouples the content and style of the foreground text, allowing the flexibility to adjust the style or content individually in the latent space.
    
    \item  QuadNet can be trained readily on real-world datasets. 
    Due to the lack of paired data in the real world, our method uses shared weight approach for training. 
    Besides, we propose effective ``cut out text areas" in the background module and adapt AdaIN in the foreground module are particularly important.
    
    \item  QuadNet generates more photo-realistic images in real scenes, and avoids the shadow residue of the original style text when editing the text content.
   
\end{itemize}
\section{Related Work}
\subsection{Scene Text Editing}
Scene Text Editing has made remarkable progress in replacing or modifying a word in the source image with another one while keeping its realistic look.
Previously, SRNet ~\cite{wu2019editing}, SwapText ~\cite{yang2020swaptext} and MG-GAN ~\cite{dendorfer2021mg} can only be trained in synthetic paired data, which may not be as effective in real-world. 
De-rendering ~\cite{shimoda2021rendering} found another way to learn the text vectorization model to get all rendering parameters, including text, position, size, font, style, special effects and hidden background, which enables to restore the background and render any text content.
TextStyleBrush ~\cite{krishnan2021textstylebrush} which is based on StyleGAN, encodes the style of a text image into a 512-dimensional style vector and then transfers it to a content image to generate the final image.
SimAN ~\cite{Luo2022SimAN} introduces a self supervised training method through similarity aware normalization. 
However, these methods are still limited to editing text content and cannot be used to change the text style at will.

Diffusion models ~\cite{sohl2015deep,NEURIPS2020_4c5bcfec,nichol2021improved} have now become a new hot topic in the field of generative models. 
Palette ~\cite{saharia2022palette} is designed based on classifier-free ~\cite{ho2021classifierfree} to perform image-to-image translation tasks, which is similar to Pix2Pix ~\cite{isola2017image}. 
We adapted Palette to perform the scene text editing task as a baseline to compare with our method.

\subsection{Scene Text Erasure}
Scene Text Erasure mainly ensures erasing the foreground text while recovering the background covered by the text. 
MTRNet ~\cite{tursun2019mtrnet,tursun2020mtrnet++} is a conditional adversarial generative network with an auxiliary mask. 
Liu et al. ~\cite{Erase2020Liu} proposes a real-world dataset called SCUT-EnsText and designs a novel GAN-based model termed EraseNet that can automatically remove text located on the natural images.
Large mask inpainting (LaMa) ~\cite{suvorov2022resolution} uses fast fourier convolutions (FFCs) to support large mask inpainting, which is used as our background inpainting module. 

\subsection{Latent Space Editing }
As image generation has progressed, many image editing methods have emerged, but most of them focus on faces or cars.
pSp ~\cite{richardson2021encoding} can directly embed images into $\mathcal{W+}$ latent space and perform editing. 
InterFaceGAN ~\cite{shen2020interpreting} found that the latent space of the well-trained generation model actually learns a disentangled representation after linear transformation.
It explores the disentangled relationship between various semantics and attempts to decouple some entangling semantics through subspace projection.
SeFa \cite{shen2021closed} is currently a relatively advanced unsupervised semantic editing method.
We apply it to SSTE tasks and compare with our QuadNet.
\begin{figure*}[t]
\centering
\begin{minipage}{0.16\linewidth}
	\centering
 Original Image
	\includegraphics[width=1\linewidth, height=0.3\linewidth]{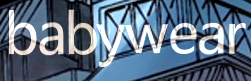}
\end{minipage}
\begin{minipage}{0.16\linewidth}
	\centering
 $-10^{\circ}$ Rotation
	\includegraphics[width=1\linewidth, height=0.3\linewidth]{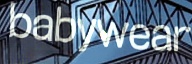}
\end{minipage}
\begin{minipage}{0.16\linewidth}
	\centering
 ${10}^{\circ}$ Rotation
	\includegraphics[width=1\linewidth, height=0.3\linewidth]{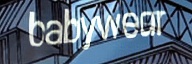}
\end{minipage}
\begin{minipage}{0.16\linewidth}
    \centering
``msyhbd" Font
    \includegraphics[width=1\linewidth, height=0.3\linewidth]{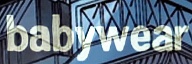}
\end{minipage}
\begin{minipage}{0.16\linewidth}
	\centering
 White Color
	\includegraphics[width=1\linewidth, height=0.3\linewidth]{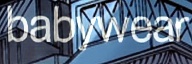}
\end{minipage}
\begin{minipage}{0.16\linewidth}
	\centering
 Green Color
	\includegraphics[width=1\linewidth, height=0.3\linewidth]{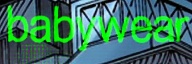}
\end{minipage}

\begin{minipage}{0.16\linewidth}
	\centering
	\includegraphics[width=1\linewidth, height=0.3\linewidth]{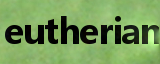}
\end{minipage}
\begin{minipage}{0.16\linewidth}
	\centering
	\includegraphics[width=1\linewidth, height=0.3\linewidth]{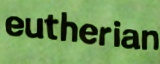}
\end{minipage}
\begin{minipage}{0.16\linewidth}
	\centering
	\includegraphics[width=1\linewidth, height=0.3\linewidth]{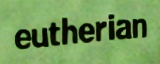}
\end{minipage}
\begin{minipage}{0.16\linewidth}
    \centering
    \includegraphics[width=1\linewidth, height=0.3\linewidth]{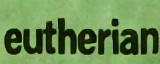}
\end{minipage}
\begin{minipage}{0.16\linewidth}
	\centering
	\includegraphics[width=1\linewidth, height=0.3\linewidth]{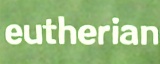}
\end{minipage}
\begin{minipage}{0.16\linewidth}
	\centering
	\includegraphics[width=1\linewidth, height=0.3\linewidth]{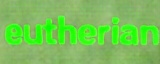}
\end{minipage}

\begin{minipage}{0.16\linewidth}
	\centering
	\includegraphics[width=1\linewidth, height=0.3\linewidth]{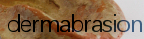}
\end{minipage}
\begin{minipage}{0.16\linewidth}
	\centering
	\includegraphics[width=1\linewidth, height=0.3\linewidth]{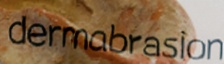}
\end{minipage}
\begin{minipage}{0.16\linewidth}
	\centering
	\includegraphics[width=1\linewidth, height=0.3\linewidth]{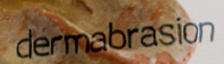}
\end{minipage}
\begin{minipage}{0.16\linewidth}
    \centering
    \includegraphics[width=1\linewidth, height=0.3\linewidth]{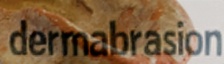}
\end{minipage}
\begin{minipage}{0.16\linewidth}
	\centering
	\includegraphics[width=1\linewidth, height=0.3\linewidth]{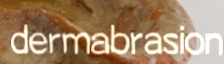}
\end{minipage}
\begin{minipage}{0.16\linewidth}
	\centering
	\includegraphics[width=1\linewidth, height=0.3\linewidth]{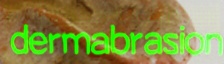}
\end{minipage}

\begin{minipage}{0.16\linewidth}
	\centering
	\includegraphics[width=1\linewidth, height=0.3\linewidth]{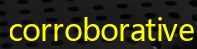}
\end{minipage}
\begin{minipage}{0.16\linewidth}
	\centering
	\includegraphics[width=1\linewidth, height=0.3\linewidth]{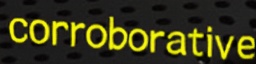}
\end{minipage}
\begin{minipage}{0.16\linewidth}
	\centering
	\includegraphics[width=1\linewidth, height=0.3\linewidth]{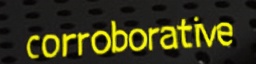}
\end{minipage}
\begin{minipage}{0.16\linewidth}
    \centering
    \includegraphics[width=1\linewidth, height=0.3\linewidth]{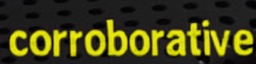}
\end{minipage}
\begin{minipage}{0.16\linewidth}
	\centering
	\includegraphics[width=1\linewidth, height=0.3\linewidth]{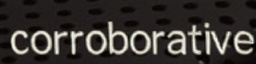}
\end{minipage}
\begin{minipage}{0.16\linewidth}
	\centering
	\includegraphics[width=1\linewidth, height=0.3\linewidth]{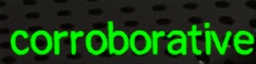}
\end{minipage}

\begin{minipage}{0.16\linewidth}
	\centering
	\includegraphics[width=1\linewidth, height=0.3\linewidth]{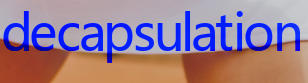}
\end{minipage}
\begin{minipage}{0.16\linewidth}
	\centering
	\includegraphics[width=1\linewidth, height=0.3\linewidth]{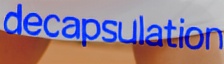}
\end{minipage}
\begin{minipage}{0.16\linewidth}
	\centering
	\includegraphics[width=1\linewidth, height=0.3\linewidth]{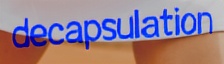}
\end{minipage}
\begin{minipage}{0.16\linewidth}
    \centering
    \includegraphics[width=1\linewidth, height=0.3\linewidth]{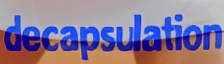}
\end{minipage}
\begin{minipage}{0.16\linewidth}
	\centering
	\includegraphics[width=1\linewidth, height=0.3\linewidth]{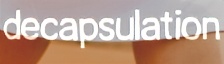}
\end{minipage}
\begin{minipage}{0.16\linewidth}
	\centering
	\includegraphics[width=1\linewidth, height=0.3\linewidth]{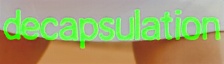}
\end{minipage}

\caption{Semantic Editing.
Column 1 is the original style text image, column 2 rotates the text by $-10^{\circ}$, column 3 rotates it by ${10}^{\circ}$, column 4 edits the text font to ``msyhbd" (Microsoft YaHei bold), column 5 changes the text color to white, and the last column changes it to green.
}
\label{fig:semanticEditing}
 \end{figure*}

\section{Methodology}
In this section, we firstly detail the model architecture, then describe the loss function and the secret of training on real-world data.
Finally we illustrate the idea of semantic editing leveraging latent space.

\subsection{Architecture}
We propose a four parts model named QuadNet to tackle the SSTE task as shown in Figure \ref{fig:architecture}. 
QuadNet considers the above mentioned background and foreground problems separately. 
Thus it is arranged as background inpainting module and foreground style editing module. 
The former adapts LaMa ~\cite{suvorov2022resolution}. 
Given an image with foreground text $I_{Ls}$ and foreground text position mask $I_{Ms}$, it erases the text pixel, recovers appropriate background with high authentic texture and outputs $G_{b}$.
The latter designs a fusion generator model similar to StyleGAN, which supports the fusion of background, foreground text style and foreground text content. 
Given two content images $T_{c1}$ and $T_{c2}$, and a style image ${I}_s$, it generates $G_{c1}$ and $G_{c2}$ whose background texture is represented by $G_{b}$.

\textbf{Background Inpainting} fine-tunes a pre-trained LaMa ~\cite{suvorov2022resolution} model to recover the background texture. 
Our goal is to inpaint the background texture in $I_{Ls}$ where is covered by the text pixel. 
Before fed to LaMa module, the mask $I_{Ms}$ is stacked with the masked image $I_{Ls}$ $\odot$ $I_{Ms}$ , resulting in a four-channel input tensor $stack($ $I_{Ls}$ $\odot$ $I_{Ms}$ $,I_{Ms}$ $)$. 
Then, we scale the image to $128 \times 2w$ uniformly, where $w$ is the image average width of the current mini batch. 
The output ${G_{b}}$ will be cropped out using a region of interest
 (RoI) align operator and deflated to the same shape as $I_{s}$ before feeding into the fusion generator.

\textbf{Style \& Content Encoders} both use ResNet34 as backbone. 
The style encoder extracts the style embedding, given $I_{s}$ denoting source style image. 
The content encoder extracts the text content representation from the content images $T_{c1}$ and $T_{c2}$, which have a gray background and the text is rendered using a fixed font. 
$T_{c1}$ and $T_{c2}$ share the same parameters of the content encoder and fusion generator. 
All images, ${I}_s$ and $T_{c1}$ and $T_{c2}$, will be scaled to $64 \times w$ uniformly. 
The outputs of content encoder and style encoder are the feature map of [$2, w/32, 512$] in shape. 
Moreover, average pooling is used to collapse the output of the style encoder into a 512D vector $\mathbf{z}$, representing the text style.

\textbf{Fusion Generator} fuses the outputs of the above three parts to generate the target image $G_{c1}$ and $G_{c2}$. 
As shown in Figure \ref{fig:architecture},
it contains the StyleMapNet, the five-layer feature pyramid generation (called FFPG) and the background fusion. 
The StyleMapNet employs a 3-layer MLP with LeakyReLU activation function. 
Its input is a 512-dimensional style vector $\mathbf{z}$, and output is a $2 \times 512 \times 5$-dimensional tensor, which is then divided into 5 vectors $\mathbf{w}^{0}, \ldots, \mathbf{w}^{4}$, spanned in [$2, 512$] , whose role is to be used for decoupling styles like InterFaceGAN ~\cite{shen2020interpreting}. 
The FFPG references StyleGAN ~\cite{karras2019style} and is designed to generate and edit foreground style text, using 5 up-sampling residual blocks. 
Each layer of the generation uses AdaIN ~\cite{huang2017arbitrary} to inject the style feature $\mathbf{w}^{i}$ during up-sampling. 
The background fusion uses a convolution with $3 \times 3$ kernel size and stride $2$ to encode the ${G_{b}}$ into a background feature map, and then concatenates it with the feature map output from the penultimate layer of the FFPG. 
Combining the background features and foreground text features, we finally obtained $G_{c1}$ and $G_{c2}$.

\subsection{Loss Function}
The loss functions aggregate L1 Loss, Perceptual Loss, Text Recognition Loss and Discriminator Loss.

\textbf{L1 Loss} measures the distortion between $I_{s}$ and $G_{c1}$ in image pixel space, where $I_{s}$ is the source style image and $G_{c1}$ is generated by the model (with the same text content as $I_{s}$). 
It can be written as:
\begin{equation}\mathcal{L}_{1}=\left\|I_{s}-G_{c1}\right\|_{1}.
\end{equation}

\textbf{Perceptual Loss} ~\cite{johnson2016perceptual} uses a pre-trained VGG19 ~\cite{simonyan2014very} model whose inputs are $G_{c1}$ and $I_{s}$. 
It measures the perceptual similarity in feature space:
\begin{equation}
\mathcal{L}_{\text {Per }}={E}\left[\sum_{i} \frac{1}{M_{i}}\left\|\phi_{i}\left(G_{c 1}\right)-\phi_{i}\left(I_{s}\right)\right\|_{1}\right],
\end{equation}
where $\phi_{i}$ represents the feature map from relu1\_1-relu5\_1 layer within the VGG19 network; $M_{i}$ is the number of elements in the feature map of the $i$-th layer.

\textbf{Text Recognition Loss} uses a pre-trained  string recognition model $\mathbf{R}$ ~\cite{baek2019wrong}, measuring the cross entropy between two character sequences:
\begin{equation}
\begin{split}
\mathcal{L}_{\text {Text}} & =\sum_{i} (\text { CrossEntropy }\left(\mathbf{R}(G_{c1_{i}}), S_{c1_{i}}\right) \\
& +\text { CrossEntropy }\left(\mathbf{R}(G_{c2_{i}}), S_{c2_{i}}\right)),
\end{split}
\end{equation}
where $\mathbf{R}(G_{c1_{i}})$, $\mathbf{R}(G_{c2_{i}})$ denotes the predicted results and $S_{c1_{i}}$, $S_{c2_{i}}$ is the text label ground truth.
The subscript $i$ means one of the batch.

\textbf{Discriminator Loss} makes generated results more realistic. 
To enhance the local features of images, our discriminator $\mathbf{D}$ adopts the structure of PatchGAN ~\cite{isola2017image}. 
We also use Spectral Normalization ~\cite{miyato2018spectral} in the discriminator, which can stabilize the training of GAN and avoid collapse mode. 
The loss is expressed as follows:
\begin{equation}\mathcal{L}_{D}=E\left(\log \mathbf{D}\left(I_{s}, T_{c1}\right)+\log \left(1-\mathbf{D}\left(G_{c1}, T_{c1}\right)\right)\right).
\end{equation}

\textbf{Total Loss}  is eventually defined as follows:
\begin{equation}
\mathcal{L}=\lambda_{1}\mathcal{L}_{1}+\lambda_{2}\mathcal{L}_{Per}+\lambda_{3} \mathcal{L}_{Text}+\lambda_{4} \mathcal{L}_{D},
\end{equation}
where $\lambda_{i}$ is the hyper-parameter. 

\subsection{Training on Real-world Data}
The above loss functions reveal the details of the shared weight approach for training on real-world data. 
The generated image $G_{c1}$ has $I_{s}$ as ground truth for various fine supervision, but $G_{c2}$ lacks labels in the real-world datasets, for $G_{c2}$ only Text Recognition Loss is used to ensure the correctness of the text content. 
The key is that $G_{c1}$ and $G_{c2}$ share some parameters in QuadNet in the generation process, so as long as $G_{c1}$ can generate images with reasonable texture and correct text content, $G_{c2}$ can as well.

In addition, the ``cut out text areas" and AdaIN are more important for training in real-world scenarios.
The background inpainting module cuts out the text area $I_{s}$ in figure $I_{Ls}$ (called ``cut out text areas"), cutting off a channel for directly copying and transferring style text of $I_{s}$ to the output.
The AdaIN used in foreground style editing module prevents the transfer of content features of $I_{s}$, solely injecting style information.
By cutting off the two direct output channels of $I_{s}$, our model is able to avoid residual text from $I_{s}$ in the generated result, and thus train well on real-world data.

\subsection{Latent Space Editing }
To tackle the SSTE task, we separate the background and foreground before embedding foreground image into latent space and perform text editing in an extended latent space.

\textbf{Embed Image into Latent Space}. 
The previous latent space editing methods embed the image in the latent space as a whole, which cannot be applied to SSTE, because the modification direction in latent space can not be found effectively when the foreground text and background texture are mixed together, and there is a high risk of damaging the background texture when editing the foreground text style or content. 
Particularly, in our task the background pixel usually has a large proportion in the image, whose quality has a huge impact on the generation performance. 

So we first use a specific module to generate a reasonable background texture, namely background inpainting module. 
Then the well-designed foreground style editing module focuses on the foreground text, embedding the foreground text into the latent space. 
It uses style encoder and content encoder to extract style embedding and content representation respectively, meaning that the content and style of the text can be modified independently. 
The text content can be changed by modifying the input of the content encoder. 
Moreover, in order to perform fine-grained adjustment of the style of foreground text, StyleMapNet is used to convert style vector $\mathbf{z}$ to layer vector $\mathbf{w}^{i}$, similar to StyleGAN ~\cite{karras2019style}. 
The latent code $\mathbf{z}$ is first transformed into an intermediate code by a non-linear mapping function, and then further transformed into $k + 1$ vectors $\mathbf{w}^{0}, \ldots, \mathbf{w}^{k}$. 
These $\mathbf{w}^{i}$ vectors were sent to the generator and the text style attributes can be edited through modifying the $\mathbf{w}^{i}$.

\textbf{Semantic Editing}. Linearly interpolating two latent codes $\mathbf{w}^{i}_{0}$ and $\mathbf{w}^{i}_{1}$, the corresponding synthesis images change continuously. 
And if $\mathbf{w}^{i}_{0}, \mathbf{w}^{i}_{1}$ are close in latent space, the corresponding images \( x_{1}, x_{2} \in \mathcal{X} \) are visually similar, where $\mathcal{X}$ stands for the image space.
Empirically, we find that for any separable text style attribute, there exists a hyperplane in latent space such that all samples from the same side are with the same attribute. 
Therefore, semantic editing can be realized based on approximation interpolation:
\begin{equation}
\mathbf{w}^{i} = \gamma \mathbf{w}^{i}_{0} + \left(1 - \gamma \right) \mathbf{w}^{i}_{1},
\end{equation}
where $\mathbf{w}^{i}_{0}$ and $\mathbf{w}^{i}_{1}$ are latent vectors from different sides of the hyperplane in latent space. 
For example, $\mathbf{w}^{i}_{0}$ stands for red color and $\mathbf{w}^{i}_{1}$ stands for blue color. 
As the value of $\gamma$ changes, $\mathbf{w}^{i}$ changes continuously between red color and blue color.

\begin{figure}[t]
\centering 
\subfigure[${I}_s$]{
\includegraphics[width=0.16\linewidth, height=0.04\linewidth]{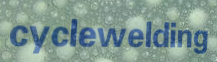}}
\subfigure[$T_{c1}$]{
\includegraphics[width=0.16\linewidth, height=0.04\linewidth]{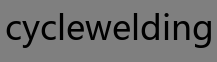}}
\subfigure[$T_{c2}$]{
\includegraphics[width=0.16\linewidth, height=0.04\linewidth]{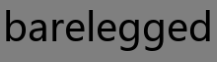}}
\subfigure[$G_{b}$]{
\includegraphics[width=0.16\linewidth, height=0.04\linewidth]{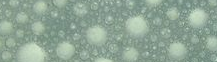}}
\subfigure[$GT_{c2}$]{
\includegraphics[width=0.16\linewidth, height=0.04\linewidth]{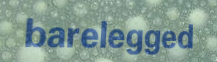}}
\caption{Sample of synthetic data.}
\label{fig.data_sync}
\end{figure}
\begin{figure}[t]
\centering
\subfigure[${I}_s$]{
\includegraphics[width=0.12\linewidth, height=0.04\linewidth]{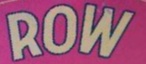}}
\subfigure[$T_{c1}$]{
\includegraphics[width=0.12\linewidth, height=0.04\linewidth]{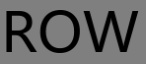}}
\subfigure[$T_{c2}$]{
\includegraphics[width=0.12\linewidth, height=0.04\linewidth]{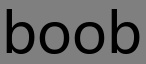}}
\subfigure[$I_{Ls}$]{
\includegraphics[width=0.19\linewidth, height=0.05\linewidth]{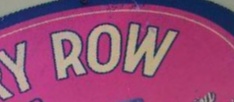}}
\subfigure[$I_{Ms}$]{
\includegraphics[width=0.19\linewidth, height=0.05\linewidth]{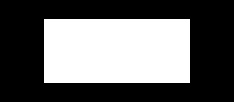}}
\caption{Sample of real-world data.}
\label{fig.data_real}
\end{figure}

    

\begin{figure}[t]
    \centering 
    \begin{minipage}{0.09\textwidth}
    \centering 
    Original
    \includegraphics[width=1\linewidth, height=0.32\linewidth]{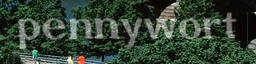}
    \end{minipage}
    \begin{minipage}{0.09\textwidth}
    \centering
    First
    \includegraphics[width=1\linewidth, height=0.32\linewidth]{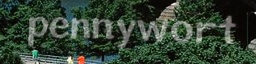}
    \end{minipage}
    \begin{minipage}{0.09\textwidth}
    \centering 
    Middle
    \includegraphics[width=1\linewidth, height=0.32\linewidth]{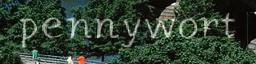}
    \end{minipage}
    \begin{minipage}{0.09\textwidth}
    \centering 
    Last
    \includegraphics[width=1\linewidth, height=0.32\linewidth]{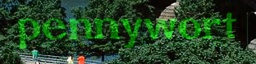}
    \end{minipage}
    \begin{minipage}{0.09\textwidth}
    \centering 
    All
    \includegraphics[width=1\linewidth, height=0.32\linewidth]{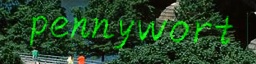}
    \end{minipage}

    \begin{minipage}{0.09\textwidth}
    \centering 
    \includegraphics[width=1\linewidth, height=0.32\linewidth]{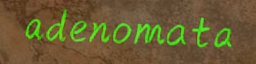}
    \end{minipage}
    \begin{minipage}{0.09\textwidth}
    \centering
    \includegraphics[width=1\linewidth, height=0.32\linewidth]{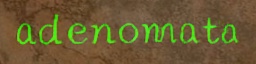}
    \end{minipage}
    \begin{minipage}{0.09\textwidth}
    \centering 
    \includegraphics[width=1\linewidth, height=0.32\linewidth]{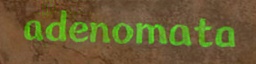}
    \end{minipage}
    \begin{minipage}{0.09\textwidth}
    \centering 
    \includegraphics[width=1\linewidth, height=0.32\linewidth]{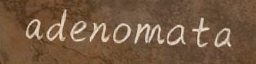}
    \end{minipage}
    \begin{minipage}{0.09\textwidth}
    \centering 
    \includegraphics[width=1\linewidth, height=0.32\linewidth]{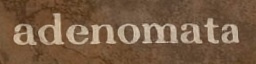}
    \end{minipage}

\caption{Swapping latent space vectors of the two original images: First ($\mathbf{w}^{0}$), Middle ($\mathbf{w}^{1}$, $\mathbf{w}^{2}$, $\mathbf{w}^{3}$), Last ($\mathbf{w}^{4}$), All ($\mathbf{w}^{0}$, \ldots, $\mathbf{w}^{4}$).
}
\label{fig:style_w_swap}
    
\end{figure}

\begin{figure*}[t]
\centering 
\subfigure[$\mathbf{w}^{0}$ about rotation]{
\label{Fig.w.rotate}
\includegraphics[width=0.3\textwidth, trim=30 10 0 50]{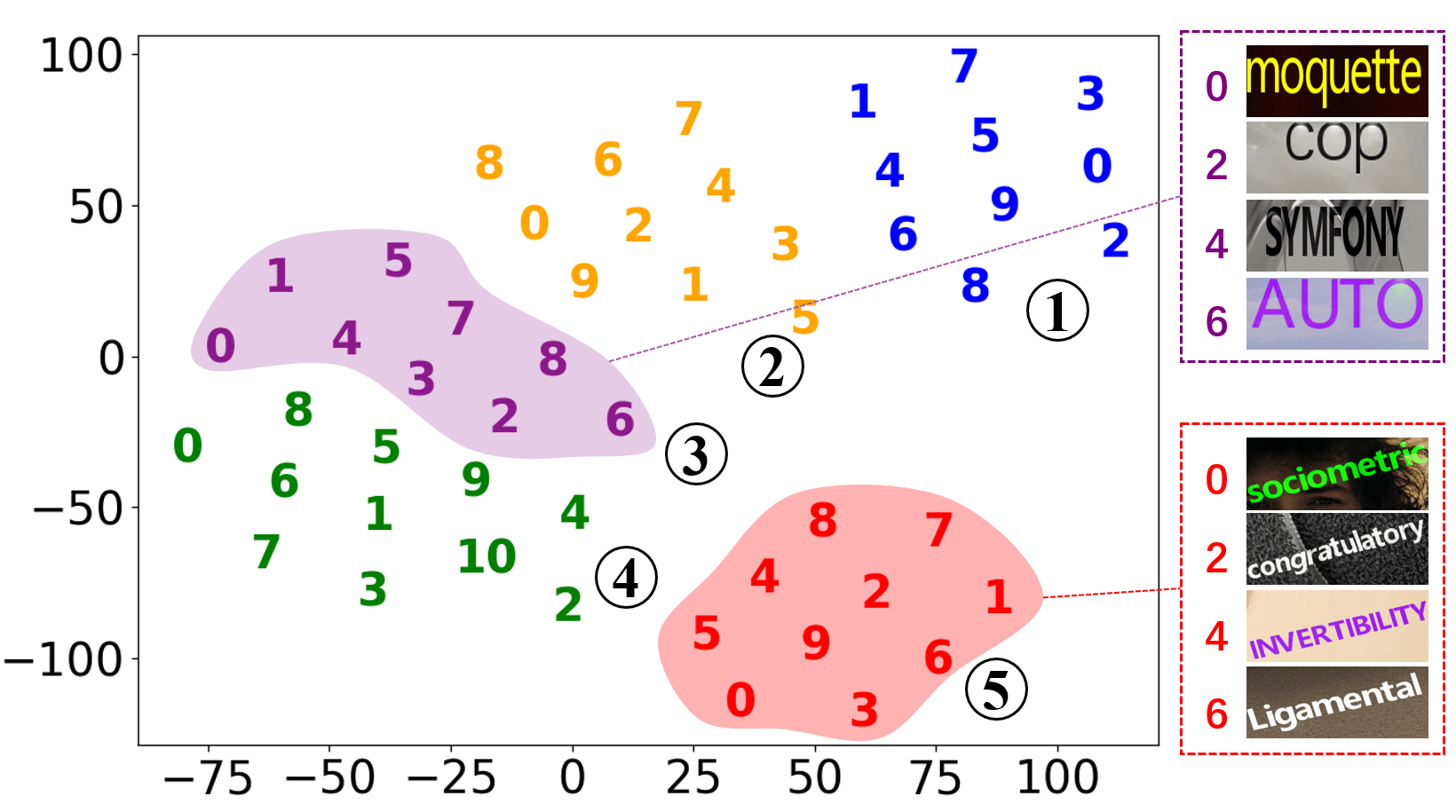}}
\subfigure[$\mathbf{{w}^{123}}$ about font]{
\label{Fig.w.font}
\includegraphics[width=0.3\textwidth, trim=20 10 0 50]{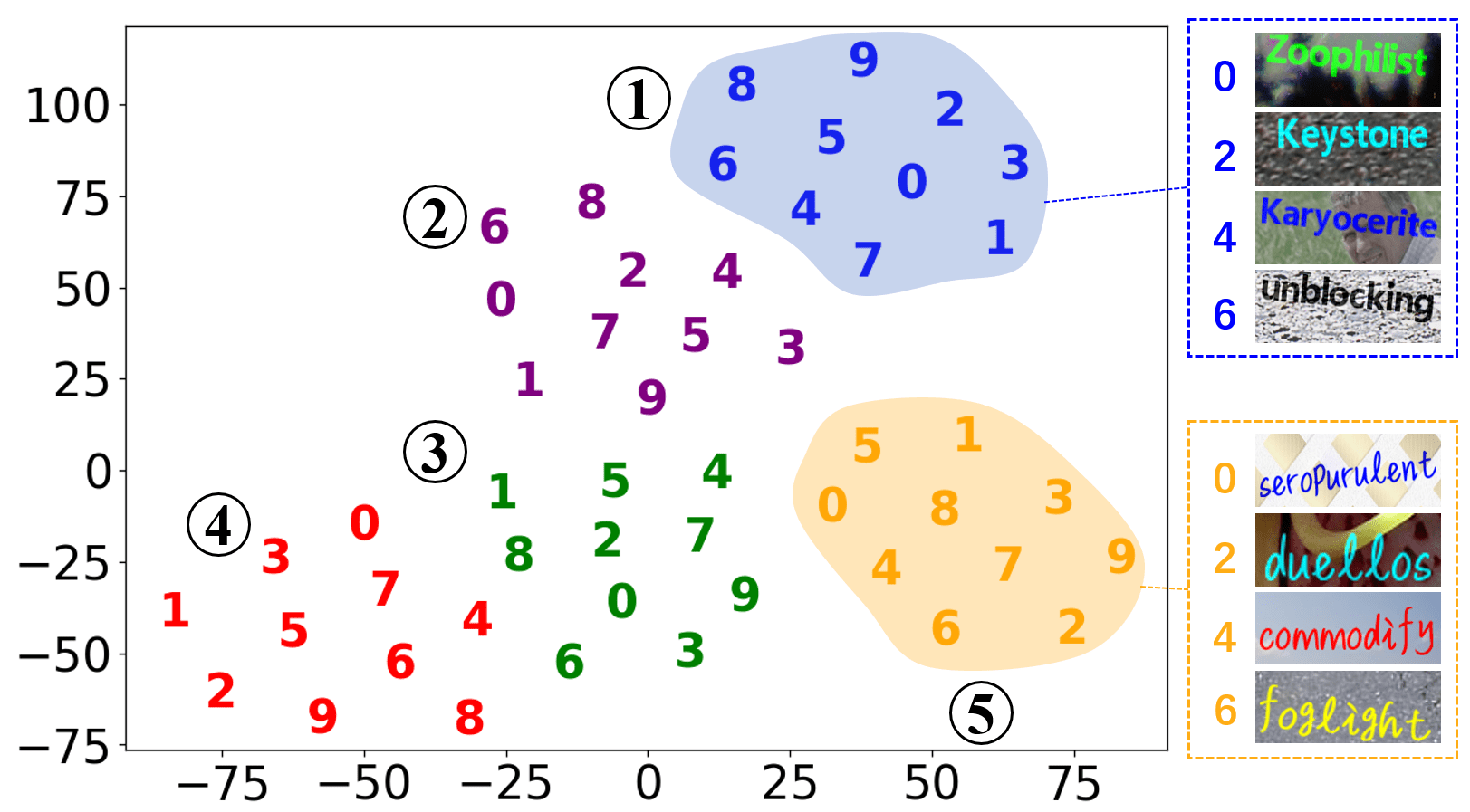}}
\subfigure[$\mathbf{w}^{4}$ about color]{
\label{Fig.w.color}
\includegraphics[width=0.3\textwidth, trim=30 10 0 50]{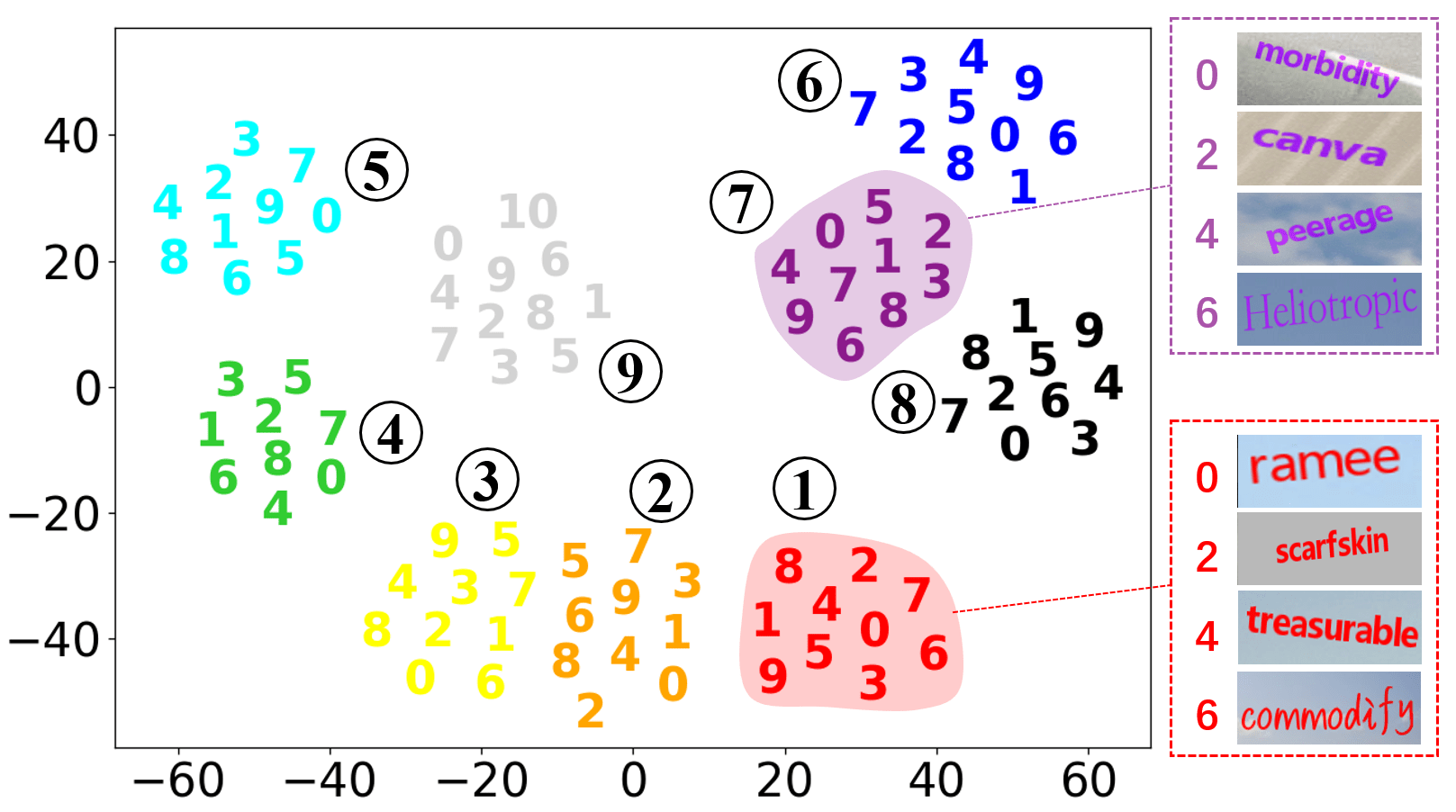}}
\caption{t-SNE visualizations of $\mathbf{w}^{i}$ about text style attributes.
a) Reduce the dimension of $\mathbf{w}^{0}$, \ding{172}-\ding{176} represent $-15^{\circ}$, $-5^{\circ}$, $0^{\circ}$, $5^{\circ}$, $15^{\circ}$.
b) Reduce the dimension of $\mathbf{w}^{123}$, \ding{172}-\ding{176} represent ``msyhbd", ``msyh", ``stkaiti", ``deng", ``handwriting".
c) Reduce the dimension of $\mathbf{w}^{4}$, \ding{172}-\ding{180} represent red, orange, yellow, green, cyan, blue, purple, white, black.}
\label{fig.W_t-SNE}
\end{figure*}

\begin{figure}[t]
\centering 
\subfigure[
Rotation angle from $10^{\circ}$ to $-10^{\circ}$: $\mathbf{w}^{0}_{0}$ and $\mathbf{w}^{0}_{1}$ represent $10^{\circ}$ and $-10^{\circ}$ respectively, $\gamma=0, 0.2, \ldots, 1.0$, $ \mathbf{w}^{0} = \gamma \mathbf{w}^{0}_{1} + \left(1 - \gamma \right) \mathbf{w}^{0}_{0}$
]{
\label{Fig.interpolation-rotate}
\includegraphics[width=0.42\textwidth, trim=0 0 0 0]{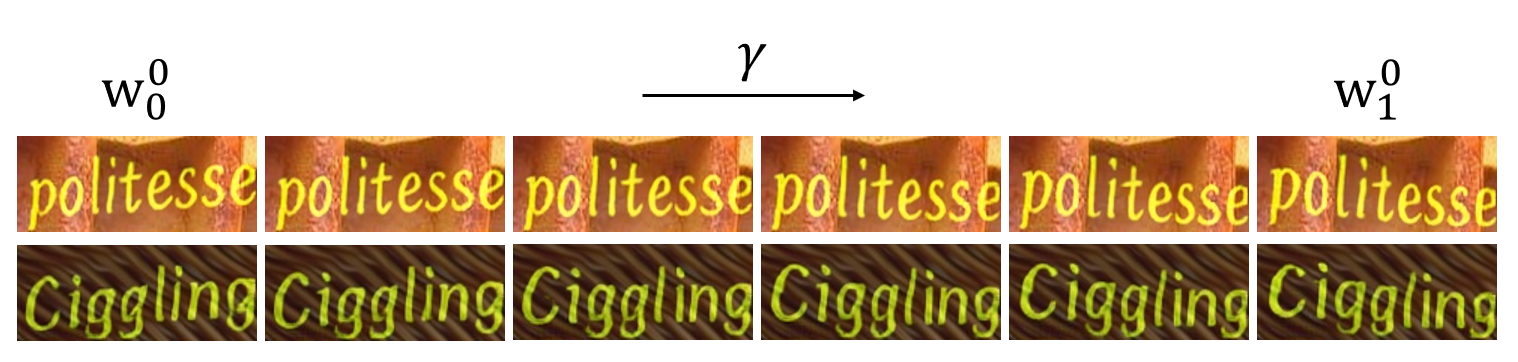}}
\subfigure[
Font from ``stkaiti" to ``deng": $\mathbf{\left[{w}^{1}_{0}, w^{2}_{0}, w^{3}_{0}\right]}$ represents ``stkaiti", $\mathbf{\left[{w}^{1}_{1}, 
w^{2}_{1}, w^{3}_{1}\right]}$ represents ``deng",
$\gamma=0, 0.2, \ldots, 1.0$, $ \mathbf{\left[{w}^{1}, w^{2}, w^{3}\right]} = \gamma \mathbf{\left[{w}^{1}_{1}, w^{2}_{1}, w^{3}_{1}\right]} + \left(1 - \gamma \right) \mathbf{\left[{w}^{1}_{0}, w^{2}_{0}, w^{3}_{0}\right]}$
]{
\label{Fig.interpolation-font}
\includegraphics[width=0.42\textwidth, trim=0 0 0 0]{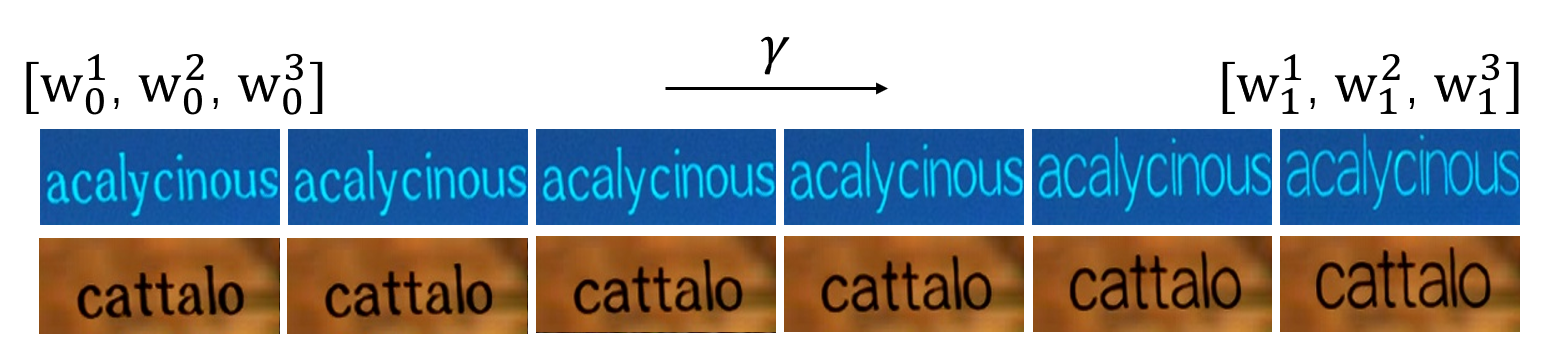}}
\subfigure[
Combine red, green and blue vector to get other colors: $\mathbf{w}^{4}_{0}$, $\mathbf{w}^{4}_{1}$ and $\mathbf{w}^{4}_{2}$ represent red, green and blue, $({\gamma}_{0}, {\gamma}_{1}, {\gamma}_{2})=(1,1,0),\ldots, (1,-1,-1)$, $\mathbf{w}^{4} = 0.5 \times ({\gamma}_{0} \mathbf{w}^{4}_{0} + {\gamma}_{1} \mathbf{w}^{4}_{1} + {\gamma}_{2} \mathbf{w}^{4}_{2}$)
]{
\label{Fig.interpolation-color}
\includegraphics[width=0.42\textwidth, trim=0 0 0 0]{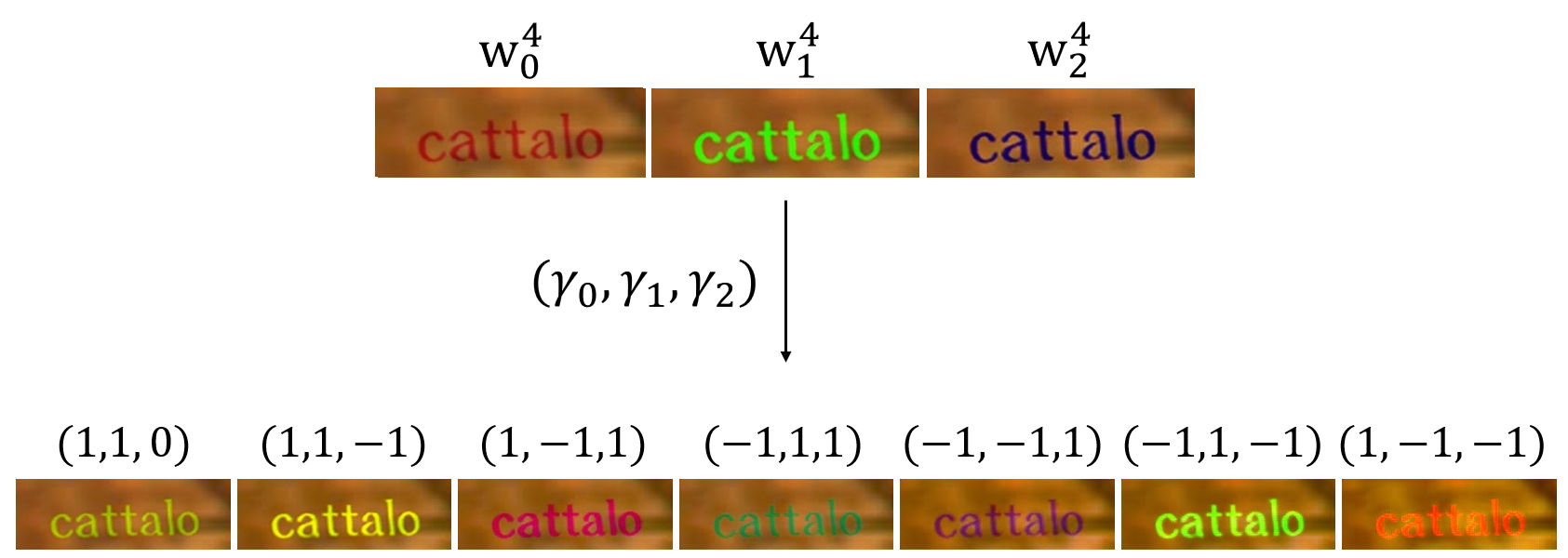}}
\caption{Linearly interpolating two latent codes $\mathbf{w}^{i}_{0}$ and $\mathbf{w}^{i}_{1}$.}
\label{fig.interpolation}
\end{figure}

\section{Experiments}
In this section\footnote{The code and dataset will be available upon acceptance, and see supplemental for more details.}, we first introduce the experimental settings, and then investigate the role of the layer vectors in the latent space, allowing us to perform semantic editing.
Finally, we carry out ablation studies and comparative experiments to evaluate the effectiveness of style text generation.

\subsection{Experimental Settings}
Our model is trained end-to-end with the help of  pre-trained models –- the recognition model ~\cite{baek2019wrong} and LaMa model ~\cite{suvorov2022resolution}. 
The model is optimized by RMSProp optimizer with a fixed learning rate of 0.0001.
We empirically set $\lambda_{1}=10, \lambda_{2} = 1, \lambda_{3}=0.1, \lambda_{4} = 1$ to balance the multiple loss terms. 
The batch size is 24 and the input images are resized to $128 \times 2w$ ($I_{Ls}$ and $I_{Ms}$) or $64 \times w$ ($T_{c1}$, $T_{c2}$, $I_{s}$ and $G_{b}$), where $w$ is the average image width of the current mini batch after scaling.
Our method is implemented using PyTorch, and the total training takes 4 days using a single RTX 3090.

\textbf{Evaluation Metrics}.  
We adopt several quantitative evaluation metrics that are commonly used in image generation, including: 
1) Frechet Inceptionthe Distance (FID) ~\cite{NIPS2017_8a1d6947}; 
2) Learned Perceptual Image Patch Similarity (LPIPS) ~\cite{zhang2018unreasonable};
3) Text Recognition Accuracy: using a text recognition engine ~\cite{baek2019wrong}, tests whether the text contents of the generated images ($G_{c2}$) are consistent with the input content images ($T_{c2}$).
See supplemental for more details.

\textbf{Synthetic Data}. 
Our model uses synthetic data for supervised learning.
We synthesized a total of 100,000 images for training and 1,000 images for ablation studies, which are similar to the data used by SRNet ~\cite{wu2019editing}. 
Figure \ref{fig.data_sync} gives an example, QuadNet takes the source style image ($I_{s}$), the target content images ($T_{c1}$ and $T_{c2}$), and the background image (${G}_b$) as input, and the output ${G}_{c2}$ is supervised using ${GT}_{c2}$.
$I_{s}$ and $GT_{c2}$ are are pairs that differ only in text content (``cyclewelding" and ``barelegged") and have the same text style and background texture. 

\textbf{Real-world Data}. We compile a real-world dataset from diverse sources, 
including SROIE, COCO ~\cite{veit2016coco}, ReCTS, ArT, LSVT, ICDAR2015 ~\cite{karatzas2015icdar}, MLT2019 ~\cite{nayef2019icdar2019} and ICDAR2019.
The training set consists of a total of 33,207 images and the test set contains 1,000 images.
One sample is illustrated in Figure \ref{fig.data_real}.
The first three images are similar to the synthetic data, but the real-world data does not have a background image (${G}_b$) and a target style text image (${GT}_{c2}$).
So we obtain a large style image (${I}_{Ls}$) and a binary image (${I}_{Ms}$) to let the model generate the background image, but ${GT}_{c2}$ is still unavailable.

\subsection{Investigation of Latent Space Editing}
\subsubsection{The Role of Image Embedding in Latent Space.}\label{sec:role_of_latent_space}
We study the semantics of the latent space vectors by swapping the layer vectors of two images. 
Figure \ref{fig:style_w_swap} shows the process of the semantic discovery.
The text in the first original image is rotated ${0}^{\circ}$ and has a print font with white color, and the second is rotated ${-5}^{\circ}$ and has a handwriting font with green color. 
We embed the two original images into the latent space respectively, and then exchange each other's layer vector $\mathbf{w}^{i}$.
We found that exchanging solely the $\mathbf{w}^{0}$ vector resulted in the two image texts swapping their rotation angles, that is, the first text is rotated at ${-5}^{\circ}$ and the second text becoming ${0}^{\circ}$.
Exchanging $\mathbf{w}^{1}$, $\mathbf{w}^{2}$, $\mathbf{w}^{3}$ together will swap their fonts, making the first image's text have a handwriting font and the second have a print font, 
Similarly, exchanging $\mathbf{w}^{4}$ will swap colors.
Furthermore, exchanging all 5 vectors will lead to superimpose effect.

To further analyze how our model successfully decouple the underlying distinct styles, we label a style text images set with 5 rotation angles, 5 fonts and 9 colors. 
We use t-SNE ~\cite{van2008visualizing} to visualize the relationship between layer vector $\mathbf{w}^{0}$ and rotation angle, $\mathbf{w}^{123}$ (merging $\mathbf{w}^{1}$, $\mathbf{w}^{2}$, and $\mathbf{w}^{3}$ into the $\mathbf{w}^{123}$ vector) and font, $\mathbf{w}^{4}$ and color, and also visualize the relationship between style vector $\mathbf{z}$ and rotation, font or color. 
The visualization results of t-SNE are shown in Figure \ref{fig.W_t-SNE}, in which the same color represents the same attribute. 
In Figure \ref{fig.W_t-SNE} (a), the image vectors $\mathbf{w}^{0}$ corresponding to the 5 text rotation angles are well distinguished, and the ${0}^{\circ}$ and ${5}^{\circ}$, ${-5}^{\circ}$ vectors are relatively close to each other, the ${15}^{\circ}$ and ${-15}^{\circ}$ vectors are relatively far apart.
Figure \ref{fig.W_t-SNE} (b) shows that layer vector $\mathbf{w}^{123}$ affect the font, and the $\mathbf{w}^{123}$ vector corresponding to different fonts are well separated.
Similarly, Figure \ref{fig.W_t-SNE} (c) shows that layer vector $\mathbf{w}^{4}$ dominates on the color. 
As a contrast, style vector $\mathbf{z}$ is not suitable to distinguish rotation angle, font or color well, see supplemental for more details.

\subsubsection{Semantic Editing in Latent Space.}
In this subsection, we demonstrate how to use latent space vectors to perform semantic editing.
We have discovered the semantics of latent space vectors, so modifying $\mathbf{w}^{i}$ vectors should allow us to fine-grained control of text style attributes, such as text rotation angle, font and color. 
Specifically, the procedure of semantic editing is as follows: 
1) label the $\mathbf{w}^{i}$, that is, use the trained model to encode image $I_{s}$ with various text style attribute labels as layer vector $\mathbf{w}^{i}$, and then the $\mathbf{w}^{i}$ also has a label, e.g. ($\mathbf{w}^{0}_{0}$, ${10}^{\circ}$), ($\mathbf{{w}^{0}_{1}}$, ${-10}^{\circ}$), ($\mathbf{\left[{w}^{1}, w^{2}, w^{3}\right]}$, ``deng" font), ($\mathbf{w}^{4}$, red); 
2) find the center of the vector space corresponding to a certain style attribute label by computing the mean of these vectors with the same label, making $\mathbf{w}^{i}$ more representative; 
3) modify layer vector $\mathbf{w}^{i}$ to achieve semantic editing by linear interpolation.

Figure \ref{fig.interpolation} illustrates the detailed process of performing semantic editing using linear interpolation.
In Figure \ref{fig.interpolation} (a), we rotate the text from $10^{\circ}$ to $-10^{\circ}$ linearly, where $\mathbf{w}^{0}_{0}$ represents $10^{\circ}$ and $\mathbf{w}^{0}_{1}$ represents $-10^{\circ}$.
By adjusting the $\gamma$ parameter, such as $\gamma=0, 0.2, \ldots, 1.0$, we will obtain multiple new $\mathbf{w}^{0}$ vectors through the formula: $ \mathbf{w}^{0} = \gamma \mathbf{w}^{0}_{1} + \left(1 - \gamma \right) \mathbf{w}^{0}_{0}$.
In Figure \ref{fig.interpolation} (b), we edit the text font from ``stkaiti" to ``deng" by altering the $\mathbf{\left[{w}^{1}, w^{2}, w^{3}\right]}$. 
In Figure \ref{fig.interpolation} (c), we mix the $\mathbf{w}^{4}_{0}$, $\mathbf{w}^{4}_{1}$, $\mathbf{w}^{4}_{2}$ vectors corresponding to the red, green, and blue colors to get more colors, such as light green, yellow, magenta, dark green, purple, bright green and orange.
By using the parameters $({\gamma}_{0}, {\gamma}_{1}, {\gamma}_{2})=(1,1,-1)$ we can get yellow. 
Figure \ref{fig:semanticEditing} displays several examples of text style semantic editing.

\begin{table}[t]
  \caption{Ablation study on synthetic data.}
  \label{Ablation-table}
  \centering
  \scalebox{0.75}{
  \begin{tabular}{lccc}
    \toprule
    Method   & Accuracy $\uparrow$ & FID $\downarrow$ & LPIPS $\downarrow$  \\
    \midrule
    w/o Background Inpainting  & \textbf{0.973}  & 76.58    & 0.33    \\
    w/o Style Encoder  & 0.847   & 66.09   & 0.31    \\
    w/o Content Encoder  & 0.000   & 96.78    & 0.38    \\
    w/o StyleMapNet  & 0.971   & 46.06    & 0.27    \\
    w/o Recognizer  & 0.949   & 51.86    & 0.28    \\
    w/o Share Weight  & 0.933  & 43.92    & 0.33    \\ \midrule
    Proposed   & 0.963 & \textbf{42.15}    & \textbf{0.26}   \\
    \bottomrule
  \end{tabular}
  }
\end{table}

\begin{table}[t]
  \caption{Ablation study on the secret of training on real-world data. Using text recognition accuracy, higher value means better effect.}
  \label{train_on_real_data}
  \centering
  \scalebox{0.75}{
  \begin{tabular}{lcc}
    \toprule
    Method   & synthetic & real-world  \\  \midrule
    DCOTA  & 0.938   & 0.002        \\
    CSAC  & 0.941   & 0.007  \\ 
    \midrule
    Proposed & \textbf{0.963} & \textbf{0.887} \\
    \bottomrule
  \end{tabular}
  }
  
\end{table}

\begin{figure}[t]
 \centering
      \begin{minipage}{0.23\linewidth}
        \centering
       Source
        \includegraphics[width=1\linewidth, height=0.3\linewidth]{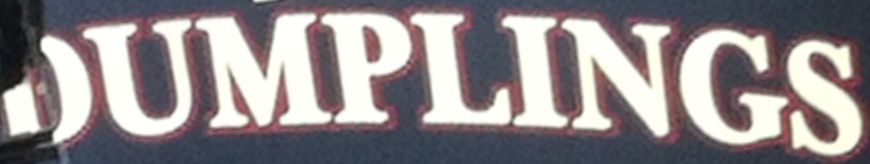} \\
        \includegraphics[width=1\linewidth, height=0.3\linewidth]{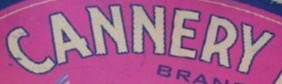} \\
        \includegraphics[width=1\linewidth, height=0.3\linewidth]{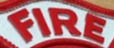} \\
        \includegraphics[width=1\linewidth, height=0.3\linewidth]{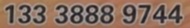} \\
      \end{minipage}
      \begin{minipage}{0.23\linewidth}
        \centering
        SRNet
        \includegraphics[width=1\linewidth, height=0.3\linewidth]{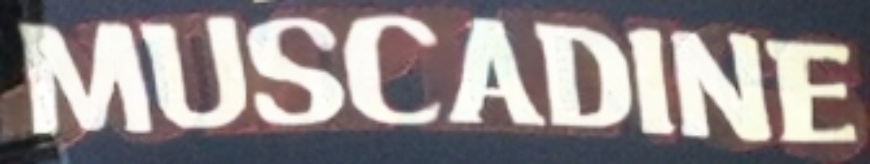} \\
        \includegraphics[width=1\linewidth, height=0.3\linewidth]{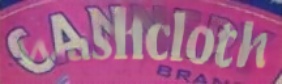} \\
        \includegraphics[width=1\linewidth, height=0.3\linewidth]{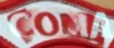} \\
        \includegraphics[width=1\linewidth, height=0.3\linewidth]{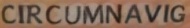} \\
      \end{minipage}
      \begin{minipage}{0.23\linewidth}
        \centering
        Palette
        \includegraphics[width=1\linewidth, height=0.3\linewidth]{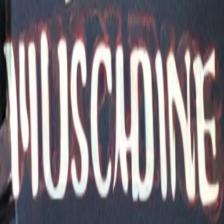} \\
        \includegraphics[width=1\linewidth, height=0.3\linewidth]{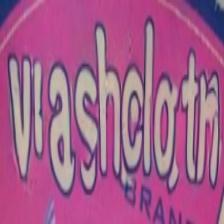} \\
        \includegraphics[width=1\linewidth, height=0.3\linewidth]{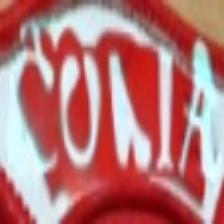} \\
        \includegraphics[width=1\linewidth, height=0.3\linewidth]{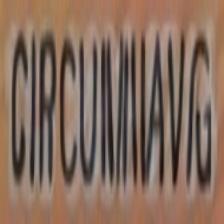} \\
      \end{minipage}
      \begin{minipage}{0.23\linewidth}
        \centering
        QuadNet
        \includegraphics[width=1\linewidth, height=0.3\linewidth]{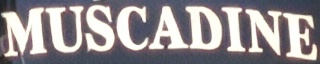} \\
         \includegraphics[width=1\linewidth, height=0.3\linewidth]{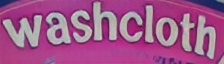} \\
         \includegraphics[width=1\linewidth, height=0.3\linewidth]{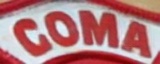} \\
         \includegraphics[width=1\linewidth, height=0.3\linewidth]{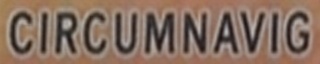} \\
      \end{minipage}

  \caption{\label{fig:artifacts} Visual comparisons on generation performance. The images generated by SRNet or Palette have residual shadows of the source image texts, while the outputs of QuadNet have no shadows.
  }
 \end{figure}

\begin{table}[t]
  \caption{Comparison of QuadNet with previous methods.}
  \label{Quantitative-table}
  \centering
  \scalebox{0.75}{
   \begin{tabular}{lccc}
  \toprule
    Method   & Accuracy $\uparrow$  & FID$^*$ $\downarrow$ & LPIPS$^*$ $\downarrow$  \\ \midrule
    SRNet ~\cite{wu2019editing}  & 0.423  & 59.05    & 0.29    \\
    Palette ~\cite{saharia2022palette}  & 0.340  & 48.10   & 0.35    \\
    \midrule
    Proposed  & \textbf{0.887}   & \textbf{37.32}    & \textbf{0.21}    \\ 
    \bottomrule
  \end{tabular}
  }
\end{table}

\subsection{Scene Text Editing Results}
In this section, we first conduct ablation studies, and then compare our results with other methods.

\subsubsection{Ablation Study.} \label{sec:ablation}
In this subsection, we analyze the role of key components of the QuadNet with quantitative results.
Table \ref{Ablation-table} and Table \ref{train_on_real_data} show the results of different settings, 
we first examine the effects of the core parts, including Background Inpainting, Style Encoder, Content Encoder, and StyleMapNet, and then further examine the text recognizer $\mathbf{R}$ and the secret of training on real-world data. 

\textbf{Separation of background and foreground}. 
After the removal of background inpainting module, the foreground and background are processed together, with the background texture also treated as a style attribute. 
From Table \ref{Ablation-table} ``w/o Background Inpainting", the recognition accuracy improves, but the FID and LPIPS both significantly decrease, which illustrates the importance of this module.

\textbf{Disentanglement of content and style}. 
Remove the Style Encoder or Content Encoder (specifically, set the output of the deleted item to a constant tensor) and then concatenate $I_s$ and $T_{c1}$ or $T_{c2}$ along the channel axis as input.
As the text content and style are coupled together, the generated result's text content cannot be well preserved.
From Table \ref{Ablation-table}, ``w/o Style Encoder" decreased recognition accuracy to 0.847, and ``w/o Content Encoder" decreased recognition accuracy to 0.
And if so, the latent space editing of text style will be difficult to achieve.

\textbf{Decomposition of text style}.
Whether or not to delete StyleMapNet has little effect on the generated results, from Table \ref{Ablation-table} ``w/o StyleMapNet".
However, according to Section \ref{sec:role_of_latent_space}, it can decompose text style and is an important structure for semantic editing.

\textbf{Benefit from recognizer}.
According to Table \ref{Ablation-table} ``w/o Recognizer", by using a text recognizer, the recognition accuracy was improved, making the generated results more readable.

\textbf{Discussion of training on real-world data}.
From Table \ref{Ablation-table} ``w/o shared weight", the shared weight approach helps to improve metrics.
Although it can be applied to training on real-world data, the ``cut out text areas" and AdaIN are more important.
We modified these two designs to demonstrate their significance.
As shown in Table \ref{train_on_real_data}, DCOTA means ``don't cut out text areas", preserve the text area in figure $I_{Ls}$, CSAC means ``concatenate style and content" rather than using AdaIN (encode style images $I_{s}$ as feature maps instead of vectors).
These two changes do not result in a significant decrease in recognition accuracy on synthetic data, but they have poor performance on real-world data (0.002 and 0.007).
On the contrary, the QuadNet's performance is 0.887.
The reason is that ${GT}_{c2}$ is unavailable in real-world, even trained with the shared weight, DCOTA and CSAC still tend to simply copy the style image $I_{s}$ as the output $G_{c2}$.

\subsubsection{Comparative Study.}
We selected SRNet ~\cite{wu2019editing} and Palette ~\cite{saharia2022palette} as the comparison methods, whose codes are open-source, and Palette has undergone some minor modifications to fit our task.
Figure \ref{fig:artifacts} illustrates the visual comparisons on inference results.
The two methods SRNet and Palette can only be trained using synthetic data, the shadow of the original style text still remains in the generated images.
Our QuadNet can be trained using a mixture of synthetic and real data.
Therefore, our method performs well on real-world data and avoids the source text shadows.
See supplemental for more qualitative samples of QuadNet.

The quantitative comparison results are shown in Table \ref{Quantitative-table}. 
Since there is no ground truth for the generated result $G_{c2}$, the metrics that can be directly measured is recognition accuracy. 
In order to better evaluate the quality of $G_{c2}$, we use it as the style image and $T_{c1}$ as the content image and perform inference again to obtain a result that can use $I_{s}$ as the ground truth, allowing us to measure the LPIPS and FID metric (without the difference between two sets of image characters). 
This iterative generation process indicates that a better LPIPS or FID score means $G_{c2}$ has a higher quality of generation.
After a simple modification, Palette performs better on the FID metric compared to SRNet, indicating the potential of diffusion models.
Our QuadNet outperforms both SRNet and Palette in all metrics remarkably. 

\section{Conclusion}
We propose QuadNet to vectorize and adjust foreground text styles in latent space. 
To our knowledge, QuadNet is the first attempt to perform fine-grained adjustment of the style of foreground text in scene text editing. 
To reduce the complexity of style editing, QuadNet performs separation of foreground text and background texture through background painting module. 
Then QuadNet does the foreground style text editing in latent space. 
To better handle real-world scenes, we developed the shared weight approach and two model designs (``cut out text areas" and AdaIN) for training.
It only needs string-level annotation information and can be trained readily on real-world datasets. 
The experiments show that our method generates better photo-realistic images, and avoids the shadow of the original style text.

\bibliographystyle{named}
\bibliography{ijcai23}

\end{document}